*Chapter 3.* Geometric Processing for Image-based 3D Object Modeling

Rongjun Qin[a,b,c,], Xu Huang[a]

[a]Department of Civil, Environmental and Geodetic Engineering

[b]Department of Electrical and Computer Engineering

[c]Translational Data Analytics Institute.

The Ohio State University

2036 Neil Avenue, Columbus, Ohio, USA

Abstract: Image-based 3D object modeling refers to the process of converting raw optical images to 3D digital representations of the objects. Very often, such models are desired to be dimensionally true, semantically labeled with photorealistic appearance (reality-based modeling). Laser scanning was deemed as the standard (and direct) way to obtaining highly accurate 3D measurements of objects, while one would have to abide the high acquisition cost and its unavailability on some of the platforms. Nowadays the image-based methods backboned by the recently developed advanced dense image matching algorithms and geo-referencing paradigms, are becoming the dominant approaches, due to its high flexibility, availability and low cost. The largely automated geometric processing of images in a 3D object reconstruction workflow, from ordered/unordered raw imagery to textured meshes, is becoming a critical part of the reality-based 3D modeling. This article summarizes the overall geometric processing workflow, with focuses on introducing the state-of-the-art methods of three major components of geometric processing: 1) geo-referencing; 2) Image dense matching 3) texture mapping. Finally, we will draw conclusions and share our outlooks of the topics discussed in this article.

## 1. Introduction

### 1.1. Background

Creating realistic 3D digital models of the physical objects and environment, termed reality-based 3D modeling, is a fundamental research theme that has been so far limiting the increasing demand of different applications, such as CFD (Computation Fluid Dynamics) (Vernay et al., 2015), 3D Geo-database (Haala and Kada, 2010), GIS (Geographic information system), urban planning, solar energy potential analysis (Freitas et al., 2015) (Strzalka et al., 2011), smart cities etc. (Biljecki et al., 2015; Gruen, 2013). This is due to the fact that the reconstruction of such models involves high-level data understanding problem, which is difficult to be reasonably addressed by the state-of-the-art computer algorithms. The existing approaches for generating city-scale and fine-grained 3D models require heavy manual involvement for measuring 3D surface, reconstructing topology (e.g. CAD models) and labeling the semantics of the objects (Flamanc et al., 2003). Reality-based modeling (RBM) is a complex task that often requires the success of a chain of important computational solutions, ranging from accurate geometric processing to high-level image and scene understanding (Gruen, 2008; Gruen et al., 2009).

Images/photographs has been the one of the predominant tools for mapping and modeling (paralleling with the LiDAR, developed since 1970s (Schwarz, 2010)). LiDAR provides direct and accurate 3D measurements of the objects, and has then been used as the standard approach for obtaining highly accurate 3D measurements of object for the past decades. Recently, the development of automated geometric processing approaches made the images again a much-favored source for 3D modeling (Remondino et al., 2014), since it is a widely available resource on almost all the platforms, ranging from personal mobile phone to satellites. 3D modeling from images is no longer expert specific: mobile applications were readily available for turning cellphone images into 3D mesh models; the boost of drone with cameras made it as a low-cost and flexible source for object modeling at fine scales. Software packages are available offering functions that could turn a set of images to 3D mesh models (Agisoft, 2017; Pix4D, 2017; Wu, 2014) with a few button-clicks. Image-based 3D point clouds are nowadays sufficient for many modeling applications (Gehrke et al., 2010; Nex and Remondino, 2014), and relevant algorithms are consistently and actively improved. In addition, images offer richer textural, spectral image boundary (Huang, 2013) information that worth more scientific exploration in aiding high-level image/data understanding and object labeling.

In general, reality-based modeling (RBM) consists of two sets of broadly defined problems, 1) Geometric processing, 2) object labeling and topology reconstruction. Geometric processing (GP) refers to the process of converting raw sensory data all the way to explicit 3D information, e.g. 3D measurements/3D triangle meshes with photo-realistic textures. The second problem, object labeling and topological reconstruction (Cornelis et al., 2008; Diakité et al., 2014; Liebelt and Schmid, 2010; Verdie et al., 2015), refers to the process of identifying the types of objects and their individual components, as well as topological relationship between different components of the object.

The geometric processing of images involves image orientation/bundle adjustment, dense image matching, point cloud meshing and texture mapping, etc.(Remondino et al., 2014) The second problem set - object labeling and topological identification, are usually operated on sparse/dense point clouds, digital surface models or triangle meshes with textures, the algorithms are rather independent of datasets, although may vary slightly with the noise levels and special characters of different sensory data.

### 1.2. Scope of this paper

The topic in this paper is image-based geometric processing under the context of the reality-based 3D modeling. Other modeling techniques that do not rely on raw data, e.g. artificial 3D objects modeling (e.g. for gaming) using rule-based methods, or procedural methods solely based shape grammars, will be out of the context. RBM is a highly disparate problem composed of many different components in both geometric aspects and high-level image & data understanding, with its solutions varying with the types and quality of data. Rather than encompass all the aspects of RBM, this paper will provide an overview on necessary techniques related to the geometric processing of images for RBM, one of the most relevant and progressive aspects in RBM in recent years. The paper will center on three main topics: 1) Image geo-referencing; 2) dense image matching and 3) texture mapping. The image/data understanding part of the RBM will not be covered, whereas relevant techniques will be partially included in approaches that incorporate image/data understanding methods for geometric processing. ***Modeling individual objects and identifying types of objects are addressed in other chapters of this book***. To outline some of the special characteristics of the image-based methods, we will partially include some technical details of data processing techniques for other types of data (e.g. LiDAR). We aim to provide an overview on the complete chain of geometric processing and the main works of each individual component, rather than a complete bibliography inclusive of all relevant works to particular subtopics. However, important/milestone works that post significant force for the geometric processing of images are expected to be included in this paper.

### 1.3. Organization of this paper

This paper aims to provide an overview of the geometric processing of images under the context of RBM. Three major components of geometric processing will be covered including: 1) Image geo-referencing; 2) Dense image matching; 3) texture mapping. We will draw our remarks and conclusions in the last section on the geometric processing aspects of RBM, and provide outlooks in further developments in this direction. This leads to sections being organized as follows: Section 2 introduces a general framework/workflow that most of the RBM taking; Section 3, 4 and 5 will discuss the aforementioned three components in detail, and the section 6 concludes this paper and draws remarks.

## 2. Reality-based Modeling – A general overview

The reality-based 3D modeling (RBM) specifically focuses on representing the physical properties of object, being the dimension, type/class and topology (Verdie et al., 2015). Initiated by the geo-community, its characteristics of being "real" drives many applications, just name a few: wide-area urban wind & flood simulation, urban planning, solar and shadow analysis etc. (Biljecki et al., 2015). The current practices in RBM in industry still rely heavily on manual intervention, and many entities have their own workflows, with variations mainly on measuring and identifying the topological structures and object elements.

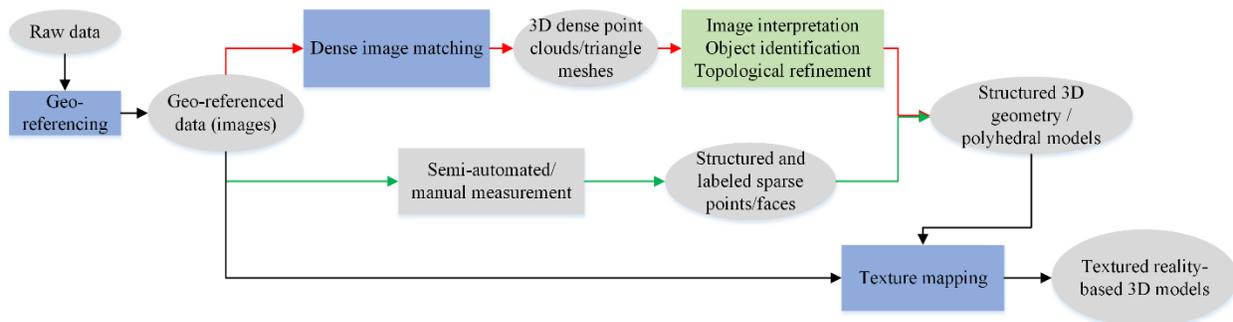

**Figure 1.** A general workflow of reality-based 3D modeling. Rectangles represent processing units and ellipses represent products (i.e. input, intermediate level results, final results). Light-blue rectangles represent the key components for geometric processing, and the light green rectangles represent the object identification and topological reconstruction. The processing chain with red arrows indicates the automated workflow and the path with green arrowed indicates the semi-automated/manual workflow.

Depending on the level of automation, the image-based RBM procedure can be broadly divided as automated and semi-automated/manual processing paths. Figure 1 illustrates common processing paths in both categories. Most of the RBM workflow starts with the geo-referencing of the raw data. The automated procedure (outlined in red arrows) aims to automate the surface reconstruction, object identification/topological reconstruction with computer algorithms. The semi-automated/manual procedure (outlined in green arrows) normally consider the 3D measurement, topological reconstruction and object/element identification as a process with operator involved with every single object being modeled. The Semi-automation are algorithms taking over part of the process including operations such as topological reconstruction (Gruen and Wang, 1998) from manually measured points or planes. With the reconstructed 3D models, the texture mapping procedure can take over by assigning & cropping the original images and assign to the relevant faces of the model (Hanush, 2010).

It should be noted the taxonomy described using automation vs. semi-automation/fully manual are rather intuitive and do not encapsulate all circumstances, and sometimes the components of two processing paths (in Figure 1, outlined by red and green arrows) can be interchangeable. For instance, the measurement of key points for topological reconstruction can be aided by densely matched point clouds. In addition, the automated methods are sometimes used as a semi-automated way: models generated by automated methods will go through a strict quality-control procedure, where manually identified errors, will be corrected either through editing tools or ad-hoc correction algorithms (Xiong et al., 2014). The improvement of RBM relies on the performance increase of each individual component (rectangles shown in Figure 1) of the workflow. Each of them has their unique challenges:

***Geo-referencing***: The geo-referencing of images refers to the process of recovering the camera interior and exterior orientation parameters. It was usually a routine process in traditional photogrammetry for ordered images, where the manual intervention were usually in measuring tie points or gross error elimination. The major challenges are usually the lack of interest/tie points in homogenous regions, and the bundle adjustment with suboptimal camera network and GCP distributions; state-of-the-art geo-referencing paradigm will be introduced in section 3.

***Dense Image matching:*** Dense image matching (DIM) refers to the processing of generating explicit 3D information (i.e. 3D point clouds, surface models). It requires per-pixel level correspondence search across stereo or multi-stereo images for accuracy 3D point determination. Although there has been much improvement for reconstructing reasonably dense 3D point clouds, the problem remains highly challenging for complex scenes with suboptimal camera networks; State-of-the-art DIM methods will be introduced in section 4.

***Texture Mapping:*** In the photogrammetry and computer vision domain, texture mapping refers to the processing of mapping realistic textures from oriented images to the generated 3D geometry. If the 3D geometry is accurate, the process of texture mapping can be performed automatically. However, there are a number challenges associated with it: 1) how to keep the textures seamless when textures are coming from different images; 2) how to deal with occluded areas; 3) how to select the best set of images from a multiple images for texture mapping. Detailed of the current practice and methods for texture mapping will be introduced in section 5.

***Object identification & topological reconstruction:*** The object identification and topological reconstruction still largely a manual process in the current industrial practice. Although there have been exciting progresses going on in terms of image understanding (Krizhevsky et al., 2012; LeCun et al., 2015), the current industrial practice still rely on fully manual/semi-automated process. The advancement in terms of geometric processing (light-blue rectangle box in Figure 1.), has evolved from a manual-guided process (tie points measurement and blunder elimination in geo-referencing and manual terrain and surface measurement) to a largely automated / fully automated process, even non-experts are able to generate   photorealistic meshes from images using the state-of-the-art commercial and open source packages (Pix4D, 2017; Wu, 2014). Therefore in the following sections, we will particularly focus on the current geometric processing methods from images, outlined by the three light-blue boxes in Figure 1.

## 3. Image Geo-referencing

Image geo-referencing refers to the process of computing the interior and exterior orientations of the camera stations in a global or local coordinate system. This normally comes along with the calibration of the interior camera parameters and lens distortions, being camera calibration. A parallel sub-field in computer vision (CV) highly relevant to geo-referencing, is referred as structure-from-motion (SFM), or pos estimation. Both CV and photogrammetry are very similar in terms of mathematical foundations, while there are differences in terms of formulations: Photogrammetry adopts algebraic operations in the scalar level for computations in Euclidean and central perspectives, while CV adopts homogeneous coordinates to represent the perspective geometry, and allows for flexibility in terms of affine distortion in the object space (Hartley and Zisserman, 2003). Moreover, photogrammetry usually uses rigorous sensors modeling for achieving the highest accuracy for mapping/3D modeling, while CV has other aims in dealing with geometry, such as visual odometry/ego-motion, or robotics, where speed and robustness are also important. More detailed comparison between these two fields is described in (Mundy, 1993; Remondino and Fraser, 2006).

In this paper, we will follow the Photogrammetry convention, while occasionally include relevant formulation/introduction in CV for comparison. In general, a geo-referencing procedure refers to a few steps shown in Figure 2. This is a general paradigm that most of the state-of-the-art geo-referencing approaches follow, with variations in methods used in individual or joints of the components.

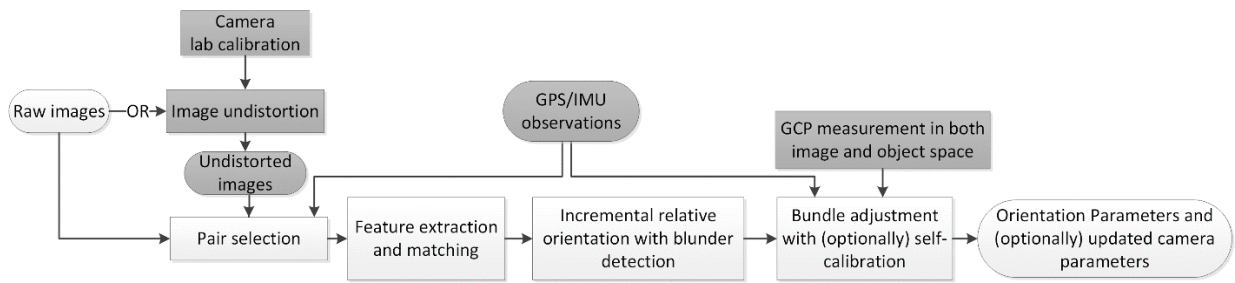

***Figure 2.***    A general workflow for geometric processing of images for 3D modeling. Rectangles represent processing units and ellipses represent inputs/outputs. Grey rectangles/ellipses are optional procedures that are case dependent.

The geometric processing workflow as shown in Figure 2 has existed and being consistently improved for decades in the field of digital photogrammetry. Nowadays it becomes a standard workflow for geo-referencing and pos estimation. The optional procedures (marked in grey in Figure 2) come along with the additionally available observations and lab calibration to improve the robustness of the workflow: Lab calibration of cameras was usually carried out in calibration field with ground control array in traditional survey based photogrammetry, while often times it may can be addressed as part of the bundle adjustment as self-calibration. GPS/IMU observations can be add-ons for the image pair selection, and/or as prior information of the orientation parameters, and/or to provide datum of the dataset. GPS/IMU observations for images are not always necessary but can be used to improve the efficiency and robustness of the workflow. The requirement of GCP for mapping purpose is usually needed to provide the datum, and they can be pose additional constraints to correct potential ground topography deformation due to lens distortions. Both the GCP and GPS/IMU, can be viewed as the observations in a general bundle solution (Gruen and Beyer, 2001).

Given a set of images, the necessary geo-referencing procedure starts with ***selecting pairs of images*** and ***extract & matching sparse correspondences*** to form an initial set of image observations, containing two or more ray points across multiple images. These image observations, often contain many blunders (erroneous observations), the ***incremental relative orientation*** aims to determine sparse corresponding points, and eliminate blunders through statistical procedures, as well as providing initial values for ***bundle adjustment***. Bundle adjustment takes into account various observations including image, GCP, GPS/IMU (approximation of exterior orientations) to accurately compute the interior and exterior parameters of images. We will provide a more detailed introduction for components illustrated above.

### 3.1.Camera Calibration

Camera calibration refers to the process of estimating camera interior orientations (focal length/ principal distance and principal points) and lens distortions. The goal of camera calibration is to separate the camera parameters such that it remains static when applying to other datasets (e.g. stereo systems), irrespective of the datasets. Calibration of metric camera mainly refers to the determination of interior orientations, as the lens distortions are normally ignorable. However, with the increasing use of consumer grader cameras for metric calculation, the camera parameters (interior orientation and lens distortions) are not dismissible. Nowadays the automated camera calibration with coded targets / natural image correspondences via bundle adjustment with self-calibration is becoming a standard procedure, particularly for close-range photogrammetry, whereas the use of 3D calibration array is no longer mandatory in this case. Nevertheless, large camera systems, such as digital airborne photogrammetric systems, or integrated multi-sensor system, still use 3D object array to estimate camera parameters, which again, follows typical least squares adjustment through the well-known collinearity equations (Schenk, 2005). The calibration parameters normally comprise the perturbation terms in the image space as $\Delta x$ and $\Delta y$ being (Fraser, 2013; Gruen and Beyer, 2001):

$$\Delta x = -\Delta x_p + \frac{\bar{x}}{c}\Delta c + \bar{x}r^2 K_1 + \bar{x}r^4 K_2 + \bar{x}r^6 K_3 + (2\bar{x}^2 + r^2)P_1 + 2P_2\bar{x}\bar{y} + b_1\bar{x} + b_2\bar{y}$$

$$\Delta y = -\Delta y_p + \frac{\bar{y}}{c}\Delta c + \bar{y}r^2 K_1 + \bar{y}r^4 K_2 + \bar{y}r^6 K_3 + (2\bar{y}^2 + r^2)P_2 + 2P_1\bar{x}\bar{y} + b_2\bar{x}$$

(1)

where $\bar{x} = (x - x_0)$ and $\bar{y} = (y - y_0)$ being the distance to the image center and $r$ being the radial distance:

$$r^2 = \bar{x}^2 + \bar{y}^2 = (x - x_0)^2 + (y - y_0)^2$$

(2)

$\Delta x$ and $\Delta y$ can be seen as the correction terms in the image space and comprised of the correction of the principal points $\Delta x_p$ and $\Delta y_p$, the correction term of principal distance $\Delta c$, the coefficients of radial distortion $K_i$, the coefficients of decentering distortion $P_i$ and the scale and affinity terms $b_1$ and $b_2$. Terms and parameter definition may vary slightly while the major concept for modeling different types of distortion remain mostly static.

Camera being able to calibrate itself in bundle adjustment does not necessarily mean it works well in all situations. Critical aspects include the camera networks (sufficient orthogonal images), coverage of image observations across the image frame, depth variation in the scene, etc. It has been well understood that there are correlations between some of the interior and exterior parameters, e.g. principal distance with respect to perspective center, principal points with respect to decentering distortion, etc. This requires the adoption of highly convergent images and orthogonal camera roll angles to break some of these correlations (Fraser, 2013). In addition, the depth variation in the scene is also crucial to accurately recover the principal distance. These critical aspects may arguably improve the calibration approaches adopted by some of the computer vision systems, where chessboard (2D scene) being the primarily calibration pattern (Zhang, 2000).

The state-of-the-art photogrammetry software systems have adopted self-calibration as a standard technique to compensate the image distortions, sometimes raised by environmental conditions (e.g. humidity, pressure, etc.). Apparently, the images used for bundle adjustment may not set forth an ideal calibration condition, which raises the probability to produce scene-dependent camera calibration parameters. Having said that, on-mission self-calibration with bundle adjustment may not be simply deemed as a black-box approach, and pre-calibration is necessary when the camera network, scene depth variation, texture richness (for interest points extraction) are not optimal, and in fact always suggested when it is possible (Remondino and Fraser, 2006).

### 3.2. Pair selection and tie point matching & extraction

Extraction and matching the tie points in the dataset is the most time-consuming part when dealing with a large image dataset. It normally starts with finding pair-wise correspondences. Selecting image pairs seems to be obvious in traditional aerial photogrammetry missions, as images are acquired following regular block structures, and most of the time the data come along with GPS/IMU observations. Interest points can nowadays be extracted and matched fully automatically, only in extreme cases tie points are needed to be added manually, such as area with poor textures, or existing tie points are unevenly distributed.

**_Pair selection:_** The use of GPS observations for pair selection in matching is rather straightforward, where neighboring images of an image defined by a pre-defined distance threshold, or *N* closest images given the GPS points (Qin et al., 2012). However, the difficulties may arise when dealing with a large set of unordered images (close-range, crowd sourcing images, and UAV (Unmanned aerial vehicles) images with no GPS headers). Exhausting all possible pairs can be order of magnitude slower (*N* times) than ordered images. One of the strategy is to perform a preliminary exhaustive matching across all the possible pairs at a much lower resolution to identify a set of cluster (Deseilligny and Clery, 2011). While it does improve computational efficiency, it may either omit important pairs due to insufficient interest points, or return too many potential pairs due to scene similarity. Methods for clustering crowd-sourcing images were developed to group images that potentially cover the same scene content (Agarwal et al., 2011; Frahm et al., 2010; Havlena and Schindler, 2014; Li et al., 2008): Iconic scene graph (Li et al., 2008) extract and cluster bag of words to build vocabulary trees (Nister and Stewenius, 2006) to cluster images, followed by geometric verification testing success of fundamental matrix estimation, and this forms a image matching graph among the unordered images. These graphs can be simplified to skeleton and additionally use parallel computations to reduce the computation time (Agarwal et al., 2011). Wu et al. (Wu, 2013) reduced the computation in the level of pair-wise correspondence search, by only matching the first 100-400 points ranked by their scale of the SIFT (Scale-invariant Feature transform) (Lowe, 2004) features for initializing the connectivity graph, which could dramatically reduce the pair-wise matching time. It may worth noting that these strategies for pair selection utilize the information content of the images, and this is at the cost of discarding pairs that do not pass initial tests, as well as reducing the pairs being used for reconstruction (skeleton of connectivity graph), with potential risks of generating highly singular camera    networks (poor convergence and overlap). Therefore, with close-range images used for generating high precision models, strategies for reducing the computation while maintaining good camera networks shall be carefully revisited.

***Feature extraction & matching:*** Feature correspondences across multiple images provide ties as the observations for recover the geometric parameters of the images. The features mostly refers to image points in practice, although there are works that incorporated lines as the observations (Habib et al., 2002). Extraction and matching are two distinctive steps: 1) a set of interest points are extracted in each individual image and 2) the matches of the points will then be carried out to generate correspondences. Overall there are mainly two types of points normally being extracted by point operators: 1) corner points, 2) blob points. Methods designed for detecting corner points usually look for points with large gradients at least in two intersected directions, as opposed to one direction (refers to lines). Examples of corner-based detectors are Moravec (Moravec, 1980), Harris (Harris and Stephens, 1988), Förstner (Förstner and Gülch, 1987), as well as their variations. Area-based point detector does not necessarily locate the point at the corners of the image, while it considers local extrema of an area in the scale space, to achieve the scale-invariance capability. Detectors in this category include operators that use difference of Gaussian (DoG), Laplacian of Gaussian (LoG) to define the interest points, such as SIFT (Lowe, 2004) and its variations such as affine-SIFT and PCA-SIFT (Ke and Sukthankar, 2004; Morel and Yu, 2009), and similar approaches such as SURF (Speed-up robust features) (Bay et al., 2006). There are also detectors that detect both types of points (Förstner et al., 2009). The detectors are generally designed towards finding scale- and rotation-invariant features, such that they appear repetitively with images contents being scaled and rotated. These detectors extract distinctive points or points with rich surrounding textures, such that these points across different images can be matched.

Feature matching often starts with two images, where every point on one images are compared with each of the point on the other image, this apparently takes $O(n_1 n_2)$ complexity, with $n_1$ and $n_2$ being the number of feature points in each of the images. Many matching methods has been proposed in the past: one class of algorithm is to compare the surrounding textures of each point such as sum of squared differences (SSD) (Kanade and Okutomi, 1994), normalized cross correlation (NCC) (Lewis, 1995), and least squares matching (LSM) (Gruen, 1985). Another set of algorithms is to extract feature vectors associated with the points as feature descriptors, and comparisons are made directly between feature descriptors. One of the most famous and milestone work was the SIFT descriptor (Lowe, 2004), where the histogram of the gradient vector at the scale space was compacted as a feature descriptor, which has pushed forward the matching correctness to a notable level, being used in many of the state of the art systems (e.g. Bundler (Snavely, 2010), visualSFM (Wu, 2014), Apero (Deseilligny and Clery, 2011) ), triggering subsequent works following similar ideas .

However, there are still many unsolved problems: lack of speed performance for feature extraction and matching is still one of the major challenges in the geometric processing pipeline. Therefore, consolidated efforts in increasing the speed performance without the cost of matching performance worth investigating. In addition, sometimes such operators generate too many points/matched on the texture-rich area (sometimes only a few pixels apart) leading to unnecessary computations, hence strategies to reduce the redundant points are also important to generate well-suited observations for geometric processing. Lastly, very often the localization of area-based point detectors are not accurate: as shown in (Remondino, 2006), the localization of SIFT point detector have an accuracy of only 2-3 pixels, that potentially creates extra uncertainties of the observation. A recommended procedure in high-precision geo-referencing by the authors, is to use corner detectors (e.g. Harris) and SIFT descriptors to perform an initial correspondence matching, and then perform a refinement matching using least square image matching to achieve high accuracy, this will be at the expense of increasing the computation complexity.

### 3.3. Incremental relative orientation with blunder detection

During a decade of development, incremental relative orientation (RO) is now becoming the standard strategy for 3D reconstruction, as well as its derived strategies (Frahm et al., 2010; Gherardi et al., 2010; Wu, 2013). The key role of incremental relative orientation is to eliminate blunders of tie points and provide initial values of the exterior orientation parameters for bundle adjustment. This is a fully automatic procedure that integrates feature correspondences across multiple images and robust statistical strategies (RANSAC (Random Sampling Consensus)). The incremental relative orientation refers to the process starting with a two-view relative orientation, followed by sequentially orienting the rest of the images given the feature correspondences. Relative orientation was developed a long time ago, while the only barrier to automation is the blunder elimination (gross errors) for wrongly matched points. Until the use of RANSAC, the major technique was to eliminate points with large residuals in relative orientation (residual-based blunder elimination), and this is very sensitive to the noise ratio of the feature matching. RANSAC used a random sampling strategy that starts with randomly sampled feature matches (observations) instead of all the observations for relative orientation (model estimation), and runs the same process for multiple times, and select the model (estimated orientation parameters) accounting for most of the observations with rseasonable residual. This has dramatically improved the automation in relative orientation and subsequently the incremental procedure, as it theoretically only requires the error rate of the matches be larger than 50%, while apparently the state-of-the-art feature extractors and matchers do much better with images in most of the applications.

The RO itself in Photogrammetry is comparatively static, where orienting two images relatively (with arbitrary datum) requires minimal five corresponding points, normally formulated as a least squares problem solved with Newton methods (Mikhail et al., 2001; Schenk, 2005). While in CV, the computation is carried out through the estimation of fundamental/essential matrix (Hartley and Zisserman, 2003), the DOF (degree of freedom) of which vary with the constraints used in the matrix, being 8-point algorithm, 5-point algorithm, 4-point and 3-point algorithms (Fraundorfer et al., 2010) (with known angles or translation). The RO problem in CV is normally solved using SVD (singular value decomposition), with additional constraints normally formed in a mathematic sense. Essentially, the flexibility of estimating fundamental matrix instead of a rigorous formulation (with unknowns represented by angles and translations) allows geometric reconstruction in an affine space. When it comes to the relative orientation for metric purpose, the rigorous formulation with iterative solutions normally delivers more accurate results and less correlated parameters.

### 3.4.Bundle adjustment

Bundle adjustment refers to the process of compute the unknowns from the collinearity equations that contain all the observations. This is a step following the incremental orientation, aiming at correcting the post estimation drift of the incremental orientation and re-adjusts the unknowns under a least squares framework. This is nowadays rather standard in photogrammetry and computer vision. In computer vision it generally refers to the minimization of the reprojection errors, being measured image points from feature correspondences (observations) as compared to the same points predicted by the bundle system (Hartley and Zisserman, 2003). In photogrammetry, the observations refer to a wider spectrum of prior information in addition to the feature correspondences, including ground control points (GCP), some of the exterior orientation parameters, or GPS/IMU observations. These observations are associated with a weight matrix accounting for the individual contributions of these observations in the bundle system. Following the formulation of (Gruen and Beyer, 2001), the observational equations can be represented as follows:

$$
\begin{aligned}
-e_B &= A_1 \boldsymbol{x_p} + A_2 \boldsymbol{t} + A_3 \boldsymbol{z} - \boldsymbol{l_B} &; \ P_B \\
-e_p &= I \boldsymbol{x_p} & - \boldsymbol{l_p} &; \ P_p \\
-e_t &= \qquad\qquad I \boldsymbol{t} & - \boldsymbol{l_t} &; \ P_t \\
-e_z &= \qquad\qquad\qquad\qquad I \boldsymbol{z} & - \boldsymbol{l_z} &; \ P_z
\end{aligned}
$$

(3)

where $A_1, A_2$ and $A_3$ being the design matrix (Jacobian Matrix) associated with the object point $\boldsymbol{x_p}$, orientation parameters $\boldsymbol{t}$, and the additional camera parameters (calibration parameters) $\boldsymbol{z}$, in (Gruen and Beyer, 2001) this refers to 11 parameter model. $e_B$, $e_p$, $e_t$ and $e_z$ are the error terms, and $\boldsymbol{l_B}$, $\boldsymbol{l_p}, \boldsymbol{l_t}$ and $\boldsymbol{l_z}$ are the observation vectors for image observations, object space point observations (GCP), orientation parameter observations and prior knowledge about the camera calibration parameters (such as from lab-based camera calibration), respectively. The weight $P_B$, $P_p, P_t, P_t$ are a priori weights measuring the confidence of the measurement, usually correlating to the standard deviation of the observations. The weight for image observations (tie points) are pretty static and this depends on the localization accuracy of the feature extractor, while it might be adjusted for tie points measured manually (for images with week connections). The second to the fourth equation of (3), corresponds to the observations directly related to the unknowns. For GCP (second equation of (3)), the weight is usually set large to enforce hard constraints to the accuracy GCPs, and it is often adjustable due to the fact that GCPs coming from different resource (directly measured GCP, or derived GCP from other sources) have different accuracy, and these can be well reflected in the a priori weight. Similar idea applies to the observations of the orientation parameters, it can either be oriented parameters for some of the images (Qin, 2014), or observations from GPS/IMU data. The observations for additional parameters (camera parameters) were normally used when the camera has been precisely calibrated. Based on similar concept, extension of the bundle adjustment with multiple camera system should be straightforward.

The design matrix ($A_1, A_2$ and $A_3$, practically will be merged as one) is a highly sparse. Past research have been carried out in optimally ordering the elements that enable efficient solvers utilizing the sparsity of the matrix, with which many cases were discussed including block-wise ordering, banded matrices (Brown, 1976), or formulation that cancel the object-space unknowns to gain efficiency (Deseilligny and Clery, 2011). Efforts in paralleling the bundle adjustment for Multi-core CPU and GPU computation are also worth-noting, which provides potential paths to real-time bundle adjustment. Additionally, efforts made initially as for relative orientation and bundle adjustment – sequential bundle adjustment (Grün, 1985), are nowadays widely used in the computer domain in visual odometry (Nistér et al., 2004). The idea of sequential adjustment is to perform bundle adjustment in the process of incremental relative orientation, while instead of starting everything over; it directly edits the element of the design/normal matrix to update the necessary elements to save the computational time. Therefore, although bundle adjustment is variable with different applications and demands, the mathematic basis remains rather statics. For well-planned data acquisition for 3D modeling, the process is largely a routine process.

### 4. Dense Image Matching

Dense image matching (DIM) refers to the process of searching for dense pixel correspondences across oriented images. Being "dense" is relative to the sparse tie points used for image orientation, often order of magnitude higher than the density of tie points (or interest points). Although not definitive, the DIM nowadays normally refers to per-pixel correspondence search (Scharstein and Szeliski, 2002). Given the image orientation, the epipolar constraints are used to reduce the 2D correspondence search problem to 1D. Despite this simplification, the DIM problem still remains challenging that attracts a great attention from the computer vision and geo-community (Scharstein and Szeliski, 2014). Depending on the number of images used in matching, DIM can be classified as stereo image matching and multi-stereo/multi-view image matching.

4.1. Stereo image matching

A stereo pair (or binocular image) refers to two images taken from different perspectives. The task is to find for each pixel in one image, its correspondence in the other. With the epipolar constraint, a popular convention is to rectify the pairs to the epipolar image space such that correspondences are in the same row between two images, thereby the matches can be represented by the column coordinate differences, termed disparity or parallax. For a match with $(y_l, x_l)$ and $(y_r, x_r)$ being the left and right pixel coordinate with a disparity $d$, it follows equation (5)

$$y_r = y_l \ , \ x_r = x_l - d \qquad\qquad (5)$$

Stereo image matching algorithms normally operate on the epipolar images with the goal to generate the pixel-wise disparity images (corresponds to either left or right view). Despite tons of solutions for this well-investigated problem, the algorithms can normally be generalized in a four –step workflow (Scharstein and Szeliski, 2002): 1) cost computation; 2) cost aggregation; 3) disparity computation; 4) disparity refinement, as shown in Figure 4-1.

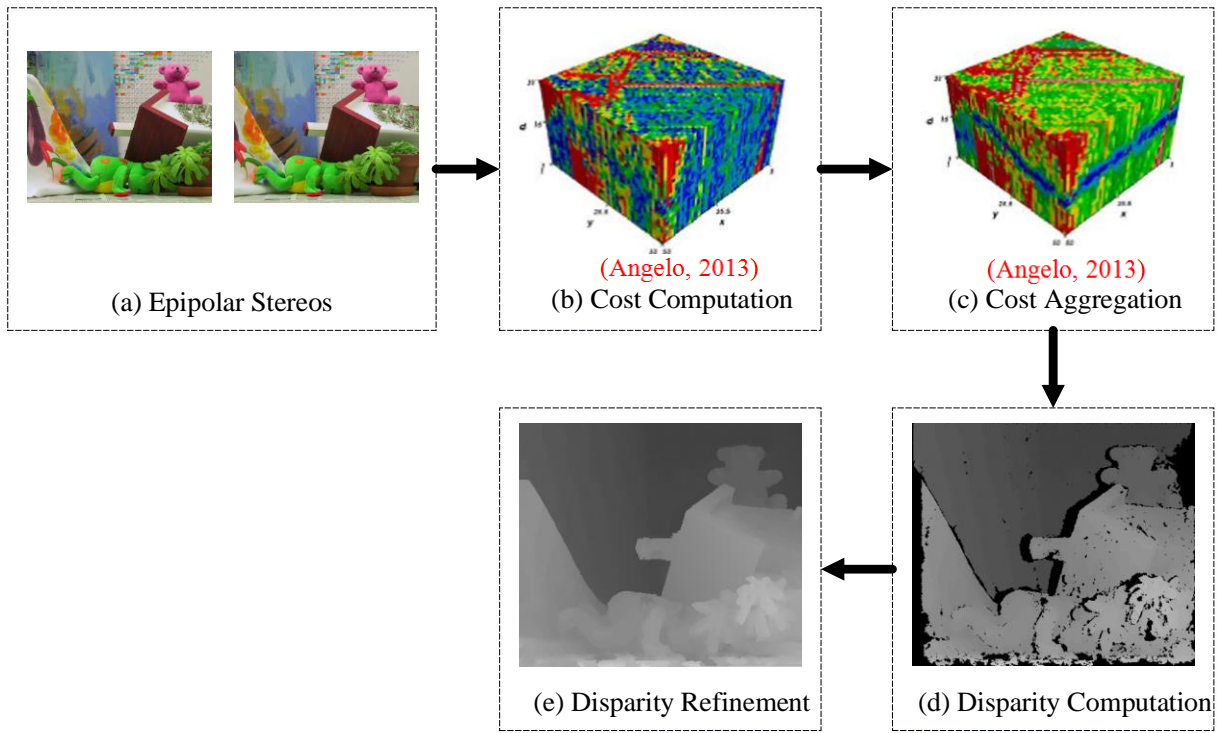

***Figure 4-1***. The workflow of Rectified Stereo Image Matching. Details are described in the text in a sequential order.

[***Read this section and see the rationale of introducing techniques-describe briefly why most algorithms can be framed within this framework, reader will be asking why you compute cost, why you need to aggregate them, why refinement is needed?***]

The stereo matching can be largely framed as a minimization problem $\underset{D}{\arg\min} E(D, I_1, I_2)$, which tends to find the optimal disparity map $D$ of the epipolar images $I_1, I_2$ by minimizing an energy function $E$. The energy is normally formulated through the measure of coherence/similarity between corresponding pixels indexed by $D$: E.g., per-pixel squared color differences aims to minimize the sum of squared error. Very often a simple per-pixel color/feature difference, also referred as cost, is not sufficient to yield accurate disparity maps [give a reference here talking about the noisy effects]. Therefore, a priori assumptions are normally posed to the disparity image itself, such that the disparity, or the 3D surface, is smooth and do not have much variation between neighboring pixels. This leads to regularizations of the $D$ in the energy function such that the neighboring pixels are inter-correlated. This is normally processed through a step called cost aggregation, being that the initial per-pixel cost computed independently are inferred and aggregated following a certain strategy to realize the regularization. The cost aggregation serves as the key to produce spatially coherence surfaces and disparity map. An optionally refinement may also be carried out to filter out or incorporate additional information to further enhance the disparity map.

### 4.1.1 Cost Computation

The cost defines the possibility of two pixels being a correct match through measuring their similarities, also called cost metric. The "cost" value being small indicates good chances of matches and vice versa. In stereo matching, the "cost" are computed for every potential matches (across a disparity range), thus generating a cost volume with a dimension of $W \times H \times L$ (Figure 4-1 (b)) $C \sim \mathbb{R}^2 \times \mathbb{R}$. Examples of the cost metrics can be either pixel-based metrics such as absolute difference (AD), the gradient differences, the insensitive measure of Birchfield and Tomasi (BT), or window-based metrics such as zero-based normalized cross correlation (ZNCC), normalized gradient (Zhou and Boulanger, 2012), Census (Zabih and Woodfill, 2005; Jiao et al., 2014; Kordelas et al, 2015) and mutual information (MI) (Paul et al, 1997). Given the large variation of radiometric properties for stereo pairs, these metrics may perform significantly different. Normally the pixel-based methods are sensitive to noises radiometric discrepancies. AD, BT or the gradient measures normally work well on images with very similar radiometry, while generally generates errors at the presence of additive or multiplitive radiometric difference. The window-based methods are normally used to account for such radiometric differences: e.g. ZNCC or NCC computes the correlation between pixel windows, which is invariant to additive radiometric differences. More advanced window-based method, such as Census and MI, are able to account for non-linear radiometric differences: Census metric adopts non-parametric transformations that utilizes the rank of the pixels within a window, and MI computes the statistical correlations between pixel windows. Such formulations naturally accounts for non-linear variations of pixel values. (Hirschmueller and Scharstein, 2009) performed a systematic studies on evaluating these cost metrics and concluded that Census and mutual information measures can achieve the best matching results under varying testing conditions. However, the drawback of these methods is that they are highly sensitive to image noises, particularly for Census metric.

More recent cost metrics adopted machine learning approaches, where large amount of sample pairs of stereo images with ground truth disparity were learned through a convolutional neural network (CNN) (Žbontar and LeCun, 2015, 2016), and applied to other images pairs to produce the confidence of matches. A preeminent fact is that among the top performed algorithms in the open source public dataset, most of them use cost metrics generated through CNN. However, a potential issue for this methods it is high demand for training data and requires large computationally resources. Moreover, its transferability to real and large-scale reconstruction still needs to be validated.

### 4.1.2 Cost Aggregation

Taking matches with the smallest cost metrics normally introduces significant errors, largely due to texture-less areas in the image and texture repetitiveness. The idea of cost aggregation is to pose smoothness constraint by assuming the scene are mostly piecewise smooth []. Such smoothness constraints can be posed either on the disparity or the normal of the planes. Algorithms using disparity smoothness constraints are called ***1D label algorithms***, since for each pixel the algorithm assigns one label. Normal vector smoothness constraints normally operates on meshes or detected plans, by assume the adjacent meshes with similar color to keep the similar normal vectors. It is also called ***3D label algorithms***, since three labels (disparity and normal direction) are assigned for every pixel.

### 4.1.2.1. 1D Label Matching

In general, 1D label matching can be classified as local matching and global/semi-global matching. Local matching assumes that all pixels with similar color in the support window should have the same disparity. They aggregate cost in a local small window. Examples of the local matching algorithms include such as bilateral filter (Yoon and Kweon, 2006), image-guided filter (He et al, 2013) and minimum spanning tree filter (Yang, 2015). These methods are weighted filters through the cost volume. Local method is simple and fast, and it normally reduces salt-and-pepper errors and the state-of-the-art filters are edge-aware, however, it easily goes into local minima of the energy function and is not generally robust in the entire region (Figure 4-3 shows speckled noises in the results).

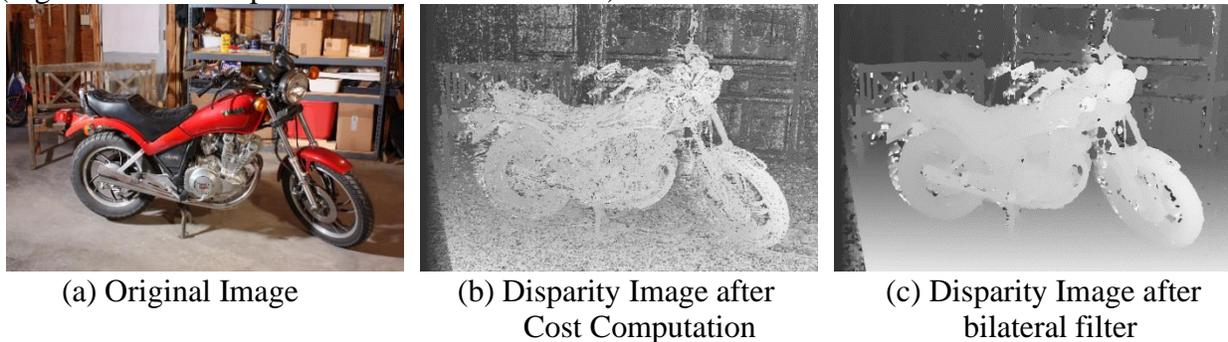

| (a) Original Image | (b) Disparity Image after Cost Computation | (c) Disparity Image after bilateral filter |

***Figure 4-3***. Disparity Images before/after Bilateral Filter.

Global/Semi-global matching methods tend to aggregate all the pixels instead of local support window. This essentially imposes a regularization term (also called smooth term) over all the pixels in the energy minimization problem:

$$\underset{D}{\arg\min} E(\boldsymbol{D}) = E_{data}(\boldsymbol{D}) + E_{smooth}(\boldsymbol{D}) \qquad (6)$$

$E_{data}$ are the cost term computed from the stereo data using cost metrics, and $E_{smooth}$ regularizes the entire disparity (rather than a support window) being smooth, wherein $E_{smooth}$ can effectively be a term that minimizes the first or even higher order derivatives of $\boldsymbol{D}$. However, this problem is a NP-hard that often requires approximate solutions. Popular solutions include graph cuts (Kolmogorov and Zabih, 2001; Papadakis and Caselles, 2010; Taniai et al., 2014), belief propagation (BP) (Yang et al., 2009), Dynamic programing (DP) [] etc. Graph cuts formulate this problem as a graph labeling problem, and belief propagation solves this problem under a general Markov Random Field model, both are iterative algorithms which tend to infer disparity consistency across the entire image, therefore they are regarded as global method. Dynamic programming operates the inferences through lines across the image grid (rather than the entire image grid), and the solution itself is not iterative, therefore it is regarded as semi-global method.

Among the existing solutions, a variant of DP method, semi-global method (SGM) [], has proven to be one of the best performing algorithm in terms its accuracy and efficiency. The idea of this method is to sequentially inferring disparities through multiple directional lines using DP (Figure 4-4 (a)). The advantage of this idea in utilizing multiple directional lines (or paths) offers compensation to single line DP/BP methods, which creates serious streak errors (comparison shown in Figure 4-4 (b-c)).

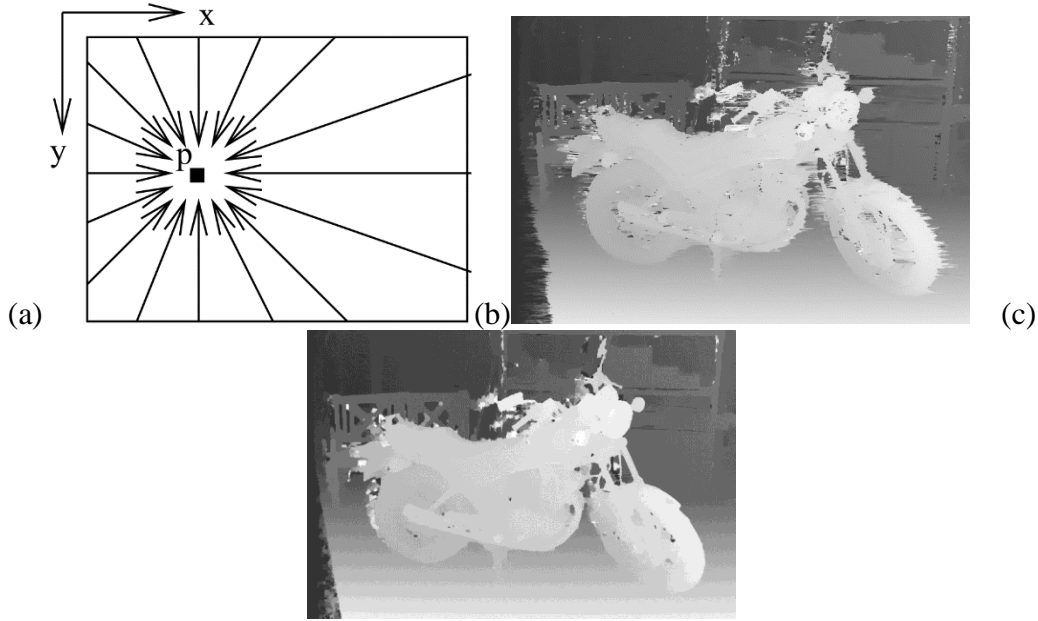

(a)       (b)         (c)

***Figure 4-4***. (a)16 Paths from all Directions (Hirschmüller, 2008), where the DP method is applied in each direction and summed up.(b) Cost Aggregation with Only One Path (c) Cost Aggregation from All Directions

Comparing to the global approaches, SGM is extremely efficient and can be applied to large format images. It can achieve the goal of quasi real-time matching with the help of GPU. The complexity of SGM is $O\,(WHL)$ (W: width, H: height, L: disparity range). Although no longer being best matching method in the middlebury test, given its trade off in efficiency and accuracy, SGM strategy are still the best performing approaches and has been widely implemented by commercial software packages such as SURE and PhotoScan, ERDAS, Smart3D, etc.

### 4.1.2.2. 3D Label Matching

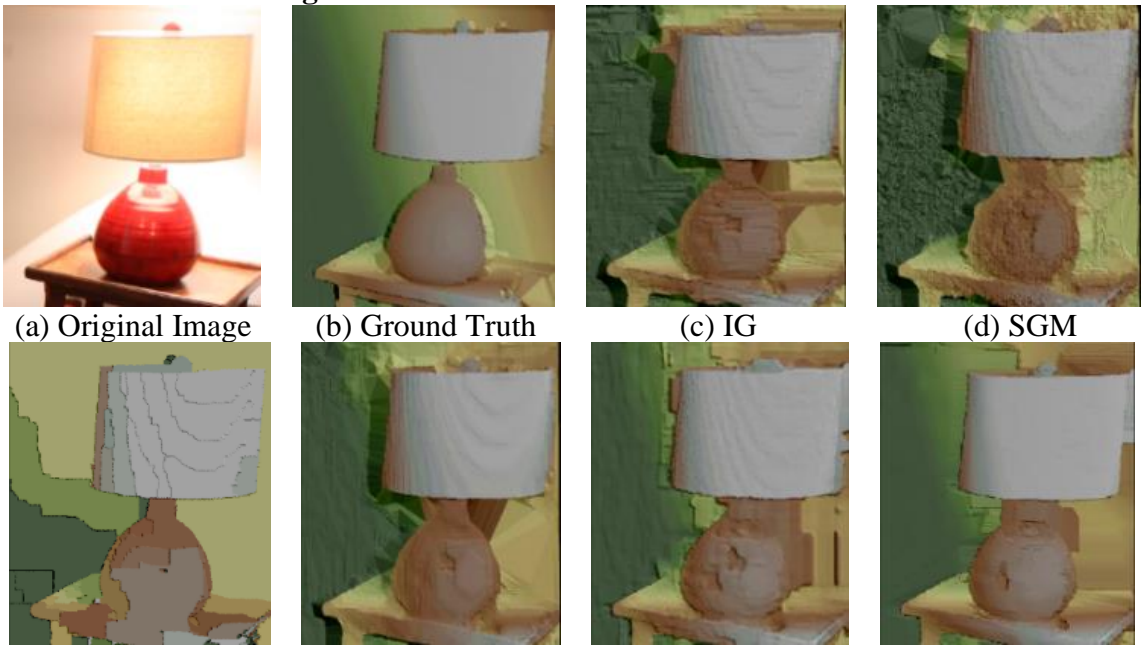

(a) Original Image  (b) Ground Truth  (c) IG    (d) SGM

(e) GC          (f) INTS          (g) NTDE          (h) PMSC

**Figure 4-6.** Results of Different Matching Methods in Slanted Planes (Huang, et al., 2017).

1D label methods assume fronto-parallel planes, meaning the algorithms work best for planes that parallel to the baseline. The ignorance of slant planes leads to the known "staircase" effects, shown in Figure 4-6 (c-g), computed by the state-of-the-art 1D labeling methods. The idea of 3D labeling methods consider the normal of the pixels in a 3D contexts, by regularizing the normal direction of neighboring pixels in addition to the disparity value (three labels, as the normal length is not considered.) (Olsson, et al., 2013). Algorithms have demonstrated that the idea of introducing the normals can effectively address this "staircase" problem (E.g., PMSC algorithm (Li, et al., 2016)) (Figure 4-7 (h)). Given the increased per-pixel unknowns, solving the Energy formulation is an ill-conditioned problem, thus almost all the 3D label algorithms assume that the scene is piecewise continuous, where normals are only computed for each segments (presuming the a segmentation is performed prior to the energy minimization), greatly reducing the number of unknowns. Considering the energy minimization problem in equation (6), the $E_{data}(\boldsymbol{D})$ that considers initial cost of matches can be constructed similarly as 1-D labeling problem (irrelevant to the normals), while the basic unit being patches/segments, e.g. the cost can be computed as the average similarity measures of pixels in the patches. The $E_{smooth}(\boldsymbol{D})$ is however different, as the normal will be taken into account: in addition to imposing smooth constraints on the disparity value, it imposes smoothness constraints to the normal direction of patches. There are normally two constraints for adjacent patches: being ***Connectivity*** and ***Coplanarity***. Connectivity panelizes disjoint surface patches in the 3D space (like disparity jump, but for the patch) (Figure 4-7 (a-b)), while Coplanarity penalizes adjacent patches with sharp angles (non-coplanar) (Figure 4-7 (c-d)).

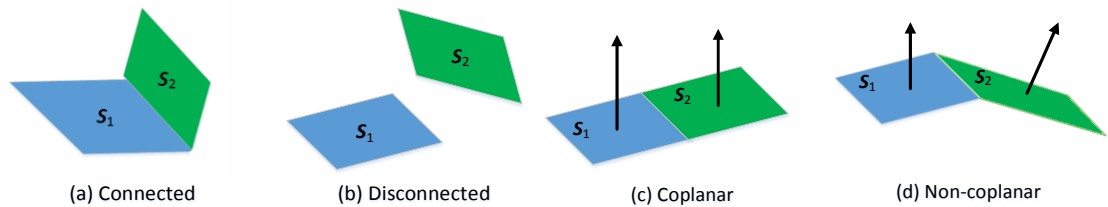

(a) Connected          (b) Disconnected          (c) Coplanar          (d) Non-coplanar

**Figure 4-7.** Patch-wise constraints.

The formulation of such constraint is straightforward [put a citation here], and some of the approximating methods used in 1-D labeling problems can be readily applied: E.g. graph cuts (Bleyer and Gelautz, 2005), belief propagation (Klaus, et al., 2006; Guney and Geiger, 2015; Yamaguchi, et al., 2012), minimum spanning tree (Veldandi, et al., 2014), least squares (Huang, et al., 2017) and scan line algorithm (Barnes, et al., 2009; Li, et al., 2016; Zhang, et al., 2015). The reconstructed surface of 3D label matching methods are normally higher in accuracy for piecewise surfaces. Given the higher complexity, the computational time is also longer than similar algorithms in 1-D labeling. Furthermore, since the methods use segmentation to generate patches, errors from segmentation (e.g. under-segmentation) may introduce additional mismatches in the final disparity.

### 4.1.3 Sub-pixel Interpolation and Consistency Check

Sub-pixel interpolation is becoming a standard processing for the disparity calculations. Normally the labeling space contains only a finite number of integers (measured by pixels). The sub-pixel interpolation tries to fit a quadratic function using disparity and its corresponding costs (or aggregated cost) to find the decimal disparity value corresponds to the valley of the quadratic function. Another common process is to use left-right consistency check to eliminate mismatches: it calculates two disparity maps by switch left and right image, and compare their consistencies. Inconsistency pixels will be eliminated, and these pixels are largely occluded pixels.

### 4.1.3 Disparity Refinements.

Many algorithms include a post-process directly on the disparity maps using image processing techniques. Such a process is often advantageous to generate clean disparity/3D surfaces. Common post-processing techniques include: 1) Speckle Filter; 2) Weight Median Filter; 3) Intensity Consistent Disparity Selection; 4) Patch-based refinement: **Speckle filter** (Hirschmüller, 2008) removes isolated disparity region that is smaller than a given threshold []; **Weight median Filter** is similar to the classic median filter, while the only difference is that it uses the intensity values to infer the ranking of the pixels to obtain edge-aware results (Mozerov and Weijer, 2015); **Intensity Consistent Disparity Selection** is a post-segmentation method that fits the best plane for each segments of the image (Hirschmüller, 2008), which is effective to correct matching errors in the textureless region. **Patch-based** method [] follows a similar strategy by defining an optimal plane for the patches, the difference is that it also allows the implementation of smooth priors to adjust neighboring planes as the 3-D labeling algorithm.

4.2.Multi-stereo/multi-view image matching

Typical mapping tasks contain much more than two overlapped images, e.g. Aerial survey, UAV (unmanned aerial Vehicles) data and close-range image. Multi-stereo/multi-view image matching methods are used to reconstruct 3D surface from a full set of overlapped images. When additional data, such as LiDAR or GIS data, matching algorithms can be customized to achieve better performances (Diebel and Sebastian, 2005; Yang, et al., 2007; Andreasson, et al., 2011; Wang and Ferrie, 2015).

4.2.1. **Multi-stereo Matching**. Multi-stereo matching is a direct extension of stereo matching by dividing multiple images into different pairs. Surfaces/depth maps are then generated for each pair using stereo matching algorithms (section 4.1), followed by a fusion step to generated the final 3D surface. In addition to the stereo matching algorithm, the critical issues of this class of method is 1) Image pair selection; 2) pair-wise fusion.

***Optimal Image Pair Selection***. Pair-selection on images with regular photogrammetric acquisition are normally easy to perform, as neighboring images can be selected given the perspective center position as if in a 2D plane [include related papers], and various parameters such as convergence and overlaps have been taken into consideration. However for unordered images, such as mobile images, or close-range, and even some irregularly flied UAV images. Optimal pair selection is important, as over-selection may significantly increase the computational time, while under-selection may results in incomplete surfaces. Assuming each images as note, a prevalent strategy is to formulate the pair selection as a graph analysis problem, where parameters of interest, such as intersection angle, baseline, resolution and repetition (Tao, 2016) etc. are set as the optimizing goal for the graph to make pair connections (Furukawa, et al., 2010).

***Pair-wise Fusion***

The fusion of the pair-wise results helps complete the final 3D surface, as well as eliminating potential blunders and improve the accuracy using redundant observations. Based the intermediate product of stereo matching, the fusion can act in meshes, point clouds or depths.

a. **Mesh Based Fusion**. Sometimes triangular meshes can be the end products of the geometric processing. This class of fusion method aims to break and merge the topology of the meshes to form a complete and non-repetitive surface mesh. (Turk and Levoy, 1994; Newcombe, et al., 2011). This includes complex mesh operations that perform triangle clipping, reconnection, as will as the consideration of eliminating spiking noises.

b. **Volumetric Fusion**. The volumetric method vote and regularize point clouds, or surface meshes in 3D volumetric spaces. Statistics of the points/meshes in each volumetric unit are to determine the optimal presence of the 3D surface (Koch, et al., 1999; Sato, et al., 2002). Such method is normally easy to implement, while the downside of it may be the potential memory demand and aliasing effects created by regular space sampling.

c. **Depth fusion**. Depth fusion is one of the most flexible class of fusion method, due to the ease of implementation. It requires a common plane that aligns the candidate depth maps, thus various image based fusion algorithm (such as bilateral filter) can be used. This method is best practiced in mapping related fusion, as the ground plane can be simply regarded the common plane for fusion [Cite SURE paper].

4.2.2. **Multi-view Matching**. Multi-view matching uses multiple images simultaneously in matching. This class of algorithm considers the evaluation of multi-ray intersection (more than two) either locally or globally. With multiple images, it is generally not possible to operate the matching process in the rectified stereo space. Instead, the matching process is performed in the 3D space, either under the representation of 3D volumetric space, or the vertical line space.

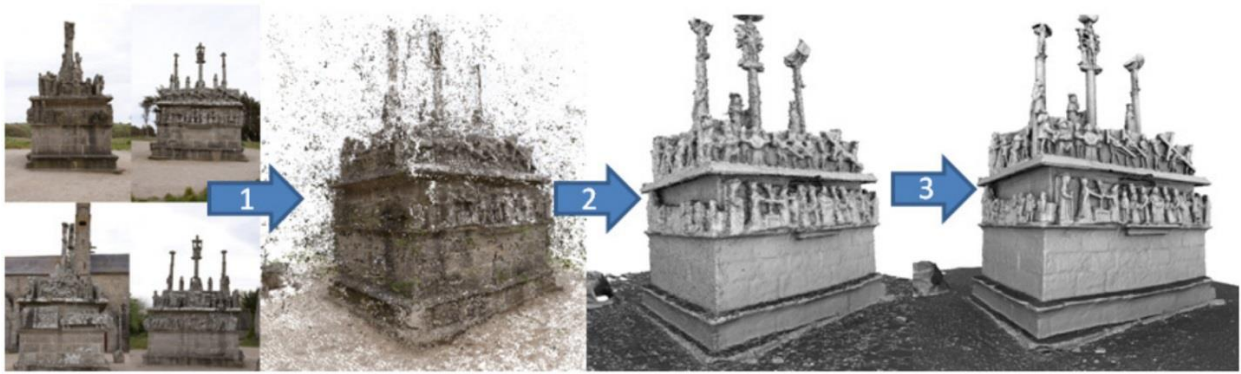

**Figure 4-14.** Reconstruction pipeline: (1) Generate a point cloud, (2) Extract a visibility consistency mesh, (3) Refine the mesh with photo-consistency optimization and regularization (Vu, et al., 2012).

a. **Quasi-dense Point Cloud Generation**. The reconstruction results of multi-view matching depend on the initial matching. If point clouds in initial matching are sparse, it is impossible to obtain a fine reconstruction result after mesh refinement. Thus, a quasi-dense point cloud is needed in multi-view matching. Most multi-view techniques generate quasi-dense point clouds locally, e.g. plane sweeping (Vu, et al., 2012), PMVS (Furukawa and Ponce, 2007), GC3 (Zhang, 2005) and self‐adaptive triangle constraint (Zhu, et al., 2010). These local matching techniques are efficient with low time complexity. It is also allowable to use multi-stereo matching methods to obtain the initial point clouds. The matching accuracies and matching completeness of multi-stereo matching are better than those of local matching. However, the time complexity of multi-stereo matching is much higher.

b. **Initial Mesh Extraction**. It consists of outlier filtering and surface reconstruction. Outliers cannot be avoided in the initial quasi-dense points. The common way is to filter outliers firstly, and then reconstruct surfaces. The research in ordered point filtering is mature. Median filter, Gaussian filter and bilateral filter (Fleishman, et al., 2003) are often adopted to remove outliers in ordered points. K-d tree filter and bounding box filter are often used to remove outliers in unordered points.

Surface reconstruction is the process of reconstructing meshes from discrete point clouds to represent the surfaces of 3D objects. The meshes should satisfy visibility constraints, namely, every mesh should be visible in at least one stereo pair. The common surface reconstruction methods include Delaunay triangulations (Amenta, et al., 2001) and Poisson surface reconstruction (Kazhdan, et al., 2006). The former one is sensitive to noises, while the later one is more robust.

Different from the above two-steps extraction, Vu, et al. achieved these two goals by relying on the Delaunay triangulation and using a visibility-based formulation to build a surface and discard outliers (Vu, et al., 2012). After extraction, these meshes are approximate to the surfaces of 3D objects, which can be refined in the following.

c. **Mesh Refinement**. It is possible to use photo-consistency constraints and smoothness constraints to refine meshes. Initial meshes can be used as the initial values of a gradient descent of an adequate energy function (Faugeras and Keriven, 1998). The energy function in Equation (16) consists of a data term and a smoothness term which correspond to photo-consistency constraints and smoothness constraints, respectively.

$$E(S) = E_{data}(S) + E_{smooth}(S)$$
$$= \sum_{i,j} \int_S v_{ij}^S(x)g(I_i, I_j)(x)dS + \int_S (\kappa_1(x)^2 + \kappa_2(x)^2)dS \qquad (16)$$

Where, $S$ is the object surface; $E_{data}$ is a data term; $E_{smooth}$ is a smoothness term; $x$ is a point of the surface; $I_i, I_j$ represent a pair of stereo; $g(I_i, I_j)(x)$ is a function of photo-consistency measure between $I_i$ and $I_j$ at $x$; $v_{ij}^S(x)$ is the function of visibility measure of $x$ in $I_i$ and $I_j$; $\kappa_1(x)$ and $\kappa_2(x)$ are principal curvatures of the surface at $x$.

The data term $E_{data}$ guarantee the photo-consistency of every point on the surface. The smoothness term penalizes strong bending between adjacent meshes. The optimal solution of Equaiton (16) is the mesh refinement results. There are several methods to get the optimal solution, e.g. variational optimization (Faugeras and Keriven, 1998; Vu, et al., 2012), least squares (Huang, et al., 2017).

3) **Vertical Line Based Matching**. The idea of vertical line based matching is similar to that of stereo image matching. It uses vertical lines to reduce the 2D correspondence matching problem into 1D, and then obtains depth images. The difference between vertical line based matching and stereo image matching is that vertical line based matching produces depth images from multi-view images instead of two overlapping images. Vertical line based matching chooses a base coordinate system (e.g. geodetic coordinate system and camera coordinate system of base images) firstly. Then, it computes a voxel-wise cost function on a 3D volume, and extract surfaces with optimization methods finally. The principle of vertical line based matching is shown in Figure 4-15.

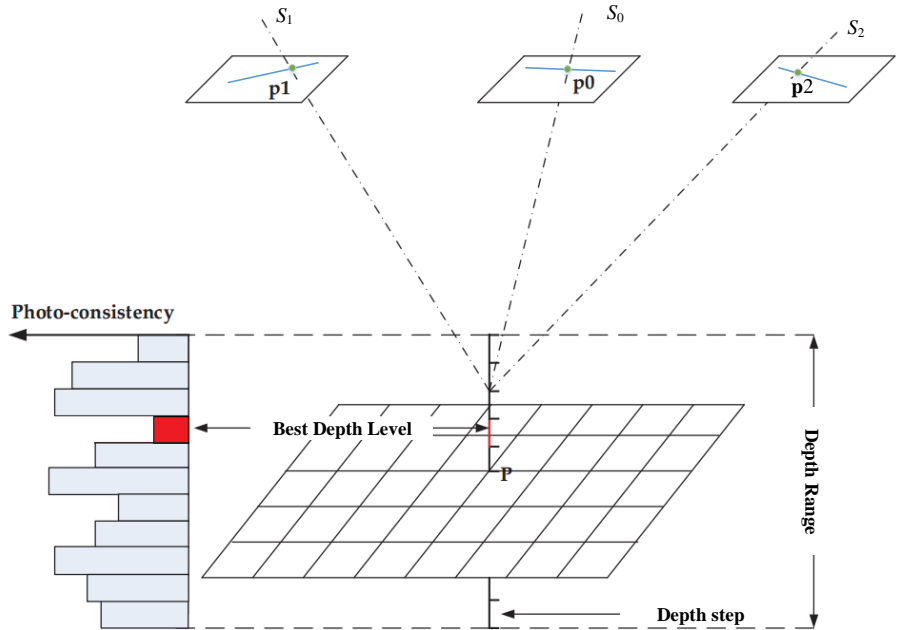

**Figure 4-15.** Principle of Vertical Line Based Matching (Zhang, et al., 2017). The depth direction is the z-axis of the base coordinate system. Depth range is pre-defined from artificialities, initial feature matching or initial DEM/DSM products. $S_0$, $S_1$ and $S_2$ are multi-view images where point **P** are visible.

In Figure 4-15, the vertical line across the grid point **P** is drawn. There are $n$ ($n = Depth\ Range/Depth\ Step$) voxels along the vertical line. If every voxel is projected onto the corresponding multi-view images, 1D tracks are generated (e.g. blue lines in Figure 4-15). The corresponding points of P must lie in the tracks. Thus, vertical lines are helpful in reducing the 2D correspondence search problem into 1D. For linear array imagery, the tracks are piece-wise continues polylines. For frame imagery, the tracks are straight lines.

Each voxel can get a photo-consistency measure (matching cost) from all the visible projected points on the images, such as $p0$, $p1$ and $p2$. Then, a cost volume can be built in the base coordinate system. The depth for each plane grid is achieved by the level which has the best photo-consistency (the lowest matching cost).

In order to avoid trapping in local minimum, most cost aggregation methods proposed in stereo image matching can also be used in vertical line based matching, e.g. graph cuts (Kolmogorov and Zabih, 2002; Vogiatzis, et al., 2005) and SGM (Zhang, et al., 2017). Vertical line based matching uses all visible images to improve the photo-consistency measures. It can get better matching results than traditional stereo matching.

Vertical line based matching can be used to produce 2.5D digital surfaces model (DSM) or real 3D models. If the former one is preferred in some applications (e.g. map surveying, roof reconstruction), geodetic coordinate system should be chosen, and then DSM can be extracted from cost volumes directly. Otherwise, fusions of multi depth images are needed to produce real 3D objects.

### 4.3. Joint geometric and semantic estimation

A recent trend tends to combine the task of retrieving object labels and recovering geometry using multiple images into a single optimization framework (Hane et al., 2013; Savinov et al., 2016). The underlying concept of this class of method is that both tasks can mutually enhance each other. This essentially becomes an extension of the volumetric method, as the object label is regarded as one of the properties of each volumetric unit. This combination has proven to be successful in achieving good geometric and semantic results (Ladický, et al., 2012; Guney and Geiger, 2015) (Yamaguchi, et al., 2014). This formulaton again, can be constructed as an energy minimization problem (Savinov et al., 2016), and the semnatic labels can be initialized using a simple image based classifier. In addition to estimate the object types, priors of the object shape can also be incorporated (Bao, et al., 2013; Wei, et al., 2014): Given training data comprised of accurate 3D models and multiple images a semantic prior can be learned to describe the general shape of the category to contrain the multi-view matching (Figure 4-16).

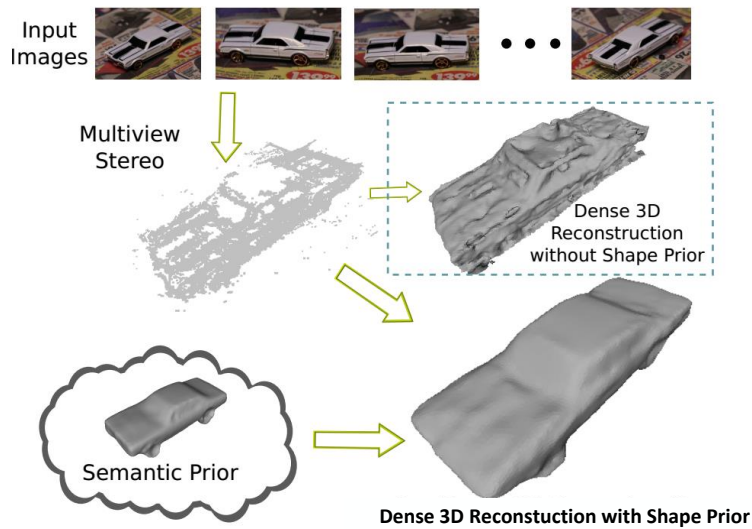

**Figure 4-16.** Semantic Prior Constrained Reconstruction (Bao, et al., 2013). 5. **Texture mapping**

Texture mapping refers to the process of assigning patches of original images to the polyhedral models and triangle meshes. It can increase the visual appearance and realism of 3D models and provide additional information for visual interpretation (e.g. patch classification and semantic labeling). In RBM, with 3D geometric models generated from oriented images, it appears to be straightforward to assign and crop the same set of oriented images to each polygonal/triangle faces (we use face thereafter) through forward projection, however the problems often rise in the following scenarios:

1) Parts of the object in 3D are occluded unwanted objects (e.g. trees occluding building façade).

2) Images from multiple views vary in lighting conditions and distance to the object of interest, leading to unbalanced color and inconsistent resolution.

3) Multi-view overlapping images result in repetitive coverage of the textures, how to select the best images for cropping the textures.

The first problem is related to object detection and texture synthesis, as the unwanted objects need to be detected and then the occluded area be replaced by synthetic textures []. In practice (engineering-grade data generation), this is often done by manual editing, or introduce more convergence images covering the objects. Most of the state-of-the-art-texture mapping solutions tend to address the problem 2) and 3). Given oriented images and triangular meshes/polyhedral models, a texture mapping procedure aims to find for each triangle/face, an optimal texture patch directly from a single image, or a fused/blended image of an optimally selected set of images. It often contain three steps:

1) **Visibility Analysis**. Test if the faces are occluded by other object, visible, or partially visible in the ray direction of the considered images, such that a list of candidate images can be identified. Details are described in Section 5.1.

2) **Image Assignment and Color Blending**. Based on the candidate images, a best set of images are selected, and colors from these images are blended in the faces. Details are described in Section 5.2.

3) **Seam Elimination**. Adjacent texture patches may come from different image sets, this creates seams in their boundaries. This step aims to eliminate these seams (Section 5.4)

**5.1. Visibility Analysis**

The visibility of a triangle/face is associated with image views. A common approach to test if a face is visible in a view is to compare its distance to the image perspective center with those of others. A face with all its points being the closet to perspective center of the view is deemed visible, or if part of its points are visible, being partially visible, otherwise this face is occluded. This is essentially the well-known Z-buffer algorithm [].

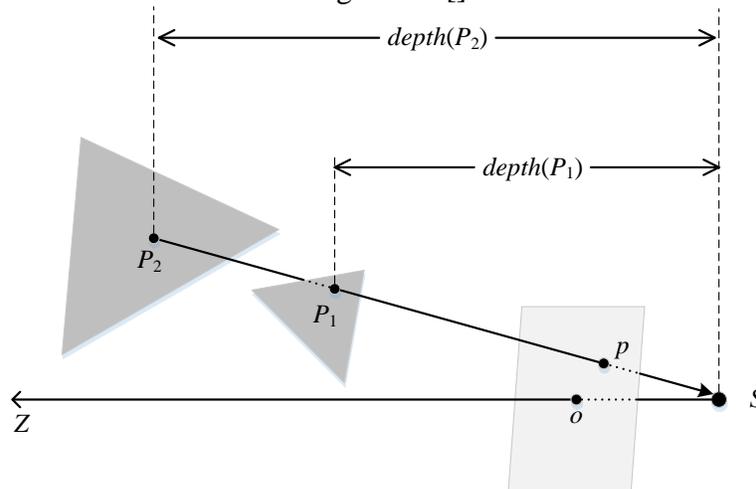

***Figure 5-6***. Projection of Meshes in Z-buffer ($P_1$ being visible). $S$ represents the camera center; $o$ represents the principal point in the image plane; $p$ represents the projective pixel; $\overrightarrow{SZ}$ represents the principal optic axis; $P_1$, $P_2$ represents 3D points in faces corresponding to the same image pixel.

In the practical implementation, testing over all the images for each face is computationally intensive, while simple intersection angle test is very effective: images with their central-ray having intersection angle with the face normal larger than 90° can be discarded, as their contribution towards the texture mapping is insignificant. By determining eligible image patches to be mapped on the face, two straightforward processes need to be implemented: 1) crop out the face footprint in the image which is beyond the image dimension, 2) invalidate occluded areas in the face footprint. The image patches in the remaining face footprint area are eligible for mapping onto the faces.

It is a slightly complicated problem to map partial textures to a face, as the mapping is performed based on the vertices. One solution is to search for other images that the face is complicated. This is normally reasonable for triangular meshes as they are typically small in units and the chances of finding images one triangle is visible is high. However, for larger polyhedral faces, this is normally solved by resynthesize a new texture patch from difference images that covers the entire image (UV mapping method, texture image shown in Figure 5-10) [].

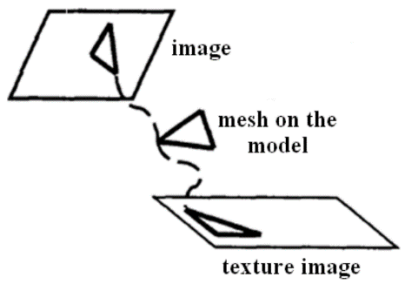

***Figure 5-10***. Texture images for mapping (Li, et al., 2010).

### 5.2. Image Assignment and Color Blending

The candidate images corresponding to a face may vary in geometric accuracy, radiance, resolutions, and their intersection angles to the face etc.. A simple image average normally give creates blurring, ghosting effects [give a citation here]. There are generally two classes of methods for image fusion and blending: 1) Weighted color blending (Li, et al., 2010; Grammatikopoulos, et al., 2007; Bernardini, et al., 2001; Callieri, et al., 2008);   2) Best image selection (Hanusch, 2009; Goldberg, 2014; Lempitsky, 2008; Niem and Broszio, 1995).

<u>***Weighted Color Blending***</u>. It weights the texture patches from different images and blending these images by linearly combining the weighted image colors. The commonly-used factors for determining the weights includes ***image distance***, ***intersection angle***, ***texture patch size***, ***color difference***, ***patch position in the original image.*** These factors and their combinations are used in different methods, and some of them are correlated: e.g. ***image distance*** and ***Texture Patch Size*** are both indicators for the texture resolution. In additional, the weighting scheme can be applied for either each patch or each ray (pixel).

***Imaging Distance*** refers to the distance from the image perspective center to the 3D face, negatively correlated with resolution, and ***texture patch size*** refers to the size of the face footprint mapped onto the image, positively correlated with resolution. And normally the higher resolution, the large weight one image will be assigned.

The smaller the ***intersection angle*** is, the smaller the distortion of the texture on the face, the larger weight will be assigned. ***Patch position in the original image*** is an easier representation of the ***intersection angle***, it measures position of the face footprint to the center of the image, the further, the larger intersection angle, the smaller weight being assigned.

The ***color difference*** measures the difference of the pixel colors to the median or mean image over all the candidate texture patches. The larger difference is, the smaller the weights are. This assumes that the color distribution of a pixel should have a mean center, where images with large differences should only be given a small weight.

***<u>Best Image Selection</u>***. Sometimes only one best image is sufficient rather than blending a number of candidate texture patches (which may introduce errors). The selection of this single best image takes into account similar indicators as the weighted color blending method, and can be either carried out per face (Local selection) (Hanusch, 2009; Goldberg, 2014), or leverage the color balancing across all the faces (Global selection) (see Figure 5-13).  The local selection optimizes a single indicator combined by various indicators as used in the weighted color blending (***image distance***, ***intersection angle***, ***texture patch size***, ***color difference***, ***patch position in the original image***). This normally generate seams between faces. The global method imposes consistencies of image selection over adjacent faces (Lempitsky, 2008; Niem and Broszio, 1995). This can be formulated as an energy minimization problem (equation (6)). Being a multiple labeling problem this can be solved via classic algorithms, e.g. Graph cut (Lempitsky, 2007) or two-steps method (Niem and Broszio, 1995).

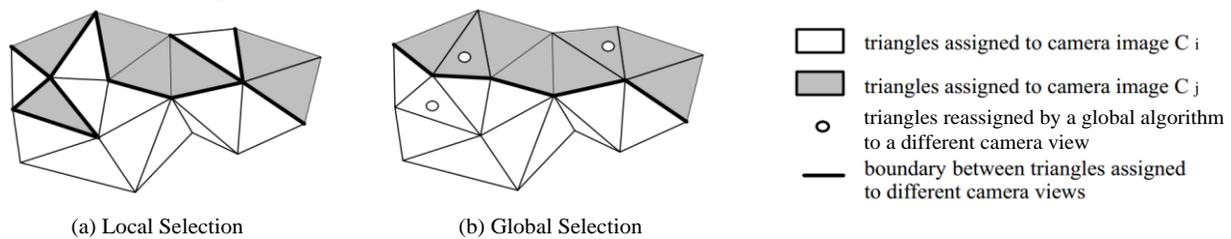

(a) Local Selection          (b) Global Selection

☐ triangles assigned to camera image C $_i$

▨ triangles assigned to camera image C $_j$

○ triangles reassigned by a global algorithm to a different camera view

— boundary between triangles assigned to different camera views

***Figure 5-13***. Image Selection after Global Method (Niem and Broszio, 1995).

### 5.3. Seam Elimination

Seams between adjacent faces, in faces with partial textures combined (also reflecting corresponding parts in the texture image) are unavoidable even through cared by the blending procedure (in section 5.2), an example is shown in Figure 5-14. Such seams not only affect the visual appearance, but also create challenges for 3D model interpretation. We introduce two representative methods that deals with seams in texture mapping.

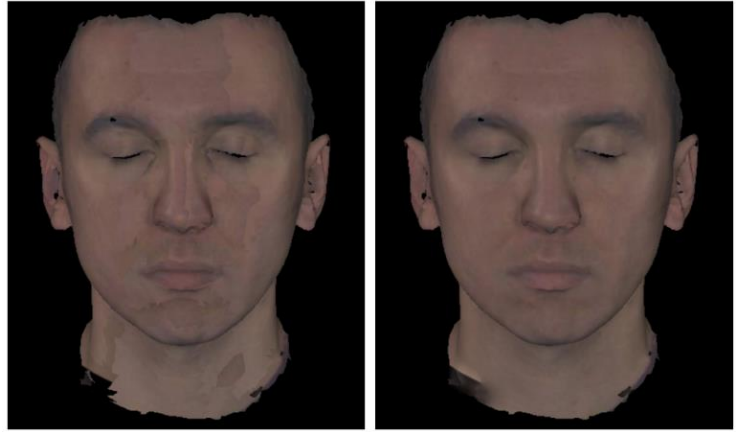

***Figure 5-14***. Before and after Seam Elimination (Velho and Sossai, 2007).

1) **Levelling Function Based Correction**.

Considering the texture space as a function, seams appears as the color discontinuities between meshes or fragments. The leveling function approach aims to construct an auxiliary function that compensates the discontinuities in the texture surfaces. The idea of the leveling functions is to compensate large color jumps by reverting the gradients in these jumps and constructing linear functions between these jumps. By adding this auxiliary function to the original function (texture colors), discontinuities can be leveled up. Figure 5-15 shows an example of this idea. The leveling function (Figure 5-15 (b)) retains the large jumps (but with a reverse direction) while keep the other corresponding part smooth.

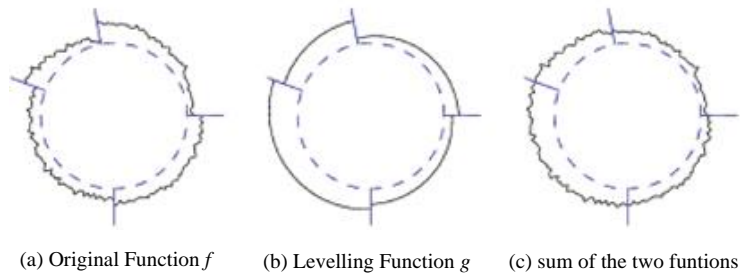

(a) Original Function $f$        (b) Levelling Function $g$        (c) sum of the two funtions

***Figure 5-15***. Seam levelling on a circumference (Lempitsky, 2008). The function values are shown as the elevations of the circumference.

2) **Brightness Correction**.

Brightness correction assumes that brightness difference is the main origin of seams. Brightness values can be computed as the lightness (L) component through color space transformation (HSL or CIELAB). The essential of brightness correction is to minimize the differences between the two L   components of adjacent faces, and correct the relative brightness values. The simplest way is to compute a constant difference between the two L components and compensate. However, the brightness difference may become complicated and errors for such correction might be propagated. Therefore this method is normally used for local correction (Hanusch, 2009).

## 5. Summary

This article provides an overview of the geometric processing of image based RBM modeling.

Particularly this includes three major topics being 1) image-based georeferencing, 2) dense image matching and 3) texture mapping. Each of the topics has been heavily investigated by scientists in the computer vision and photogrammetry community. Although the geometric processing techniques can be largely automated nowadays, particularly for regularly acquired images, our review of the methods still reveal the fact that the components that requires intelligences, for example, feature point extraction, dense corresponding search, and unwanted object removal in texture mapping, is still data and scene. The widely used piecewise linear assumption is not sufficient to cover all types of complex objects in reality, and this normally fails on fence-like objects or trees structures. Moreover, methods on critical scenarios such as reflecting surfaces, transparent surface that practically work are still lacking. Across the whole processing, there is an overall lack of learning-based methods that account for such critical situations. With the prevalent machine learning techniques applied in various domains in computer vision and photogrammetry, we foresee that more and more efforts using machine learning approaches will be casted in solving these problems.

## 5. Summary

This is a pre-publication version of the published book chapter in 3D/4D City Modelling: from sensors to applications. The final contents are subject to minor edits

This is a pre-publication version of the published book chapter in 3D/4D City Modelling: from sensors to applications. The final contents are subject to minor edits